\documentclass[sigconf,nonacm]{aamas} 
\usepackage{balance} 
\usepackage[utf8]{inputenc} 
\usepackage[T1]{fontenc}    
\usepackage{hyperref}       
\usepackage{url}            
\usepackage{booktabs}       
\usepackage{amsfonts}       
\usepackage{nicefrac}       
\usepackage{microtype}      
\usepackage{xcolor}         
\usepackage{graphicx} 

\usepackage{tabularx} 
\usepackage{svg}
\usepackage{placeins}
\usepackage{multirow}
\usepackage{tikz,pgfplots}

\usepackage{amsmath}

\title[Finding the Weakest Link]{Finding the Weakest Link: Adversarial Attack against Multi-Agent Communications}

\author{Maxwell Standen}
\affiliation{
  \institution{The University of Adelaide}
  \city{Adelaide}
  \country{Australia}} 
\affiliation{
  \institution{DST Group}
  \country{Australia}
  }
\email{maxwell.standen@adelaide.edu.au}
\email{max.standen1@defence.gov.au}

\author{Junae Kim}
\affiliation{
  \institution{DST Group}
  \country{Australia}}
\email{junae.kim@defence.gov.au}

\author{Claudia Szabo}
\affiliation{
  \institution{The University of Adelaide}
  \city{Adelaide}
  \country{Australia}}
\email{claudia.szabo@adelaide.edu.au}

\begin{abstract}

Multi-agent systems rely on communication for information sharing and action coordination, which exposes a vulnerability to attacks. We investigate single-victim communication perturbation attacks against Multi-Agent Reinforcement Learning-trained systems and propose methods that use gradient information from the Jacobian to identify which messages, agent, and timesteps are most susceptible to attack and have the greatest impact on the system. We enhance these methods with two proposed adversarial loss functions that trade-off attack success for attack impact which also create more effective perturbations. We empirically demonstrate the effectiveness of our methods against two different multi-agent communication methods in navigation, PredatorPrey, and TrafficJunction environments. Our results show that our novel message selection method achieves a similar or greater impact than random message selection across almost all tested scenarios. Our victim selection, message selection, tempo, and loss functions improve attack effectiveness in half of the thirty scenarios we tested.
\end{abstract}


\newcommand{\BibTeX}{\rm B\kern-.05em{\sc i\kern-.025em b}\kern-.08em\TeX}

\DeclareUnicodeCharacter{2212}{$-$}
\DeclareUnicodeCharacter{0394}{$\Delta$}
\DeclareUnicodeCharacter{03B5}{$\epsilon$}
\DeclareMathOperator{\mean}{mean}
\DeclareMathOperator{\median}{median}
\DeclareMathOperator{\topk}{top}

\begin{document}




\maketitle 


\section{Introduction}
Multi-agent systems have promising applications in a range of important functions such as cyber security \cite{tolba_multi-agent_2024, finistrella_multi-agent_2024}, but require communication for coordination and information sharing to operate in complex partially-observable environments \cite{oliehoek_concise_2016}. Machine-learnt communication protocols have been demonstrated to increase bandwidth efficiency \cite{foerster_learning_2016-1, zhang_succinct_2020} and noise tolerance \cite{foerster_learning_2016-1, freed_communication_2020} and thus have the potential to increase the speed and effectiveness of decision making. As learnt communication mechanisms are applied to real-world problems, it is paramount that the risks of such communication are well understood. Effective attacks allow these risks to be understood and new mitigations to be developed. However, existing attacks \cite{tu_adversarial_2021, sun_certifiably_2023, xue_mis-spoke_2022, ma_grey-box_2023} are inefficient because they do not target vulnerable messages and timesteps. Our work proposes novel methods that address these key gaps to enhance the effectiveness of attacks and improve our understanding of the vulnerability of multi-agent communications.

When evaluating the robustness of multi-agent communication, the twin goals of an attacker, namely, affecting a target system and avoiding detection, need to be considered. To affect a multi-agent system, adversarial attacks may intercept and perturb messages between agents. These communication perturbation attacks are a nascent topic of research \cite{tu_adversarial_2021, sun_certifiably_2023, xue_mis-spoke_2022, ma_grey-box_2023} and there remain unaddressed aspects of these attacks that can improve their effectiveness, namely, which messages to perturb and when to attack. 
Previous attacks arbitrarily select which messages will be perturbed \cite{tu_adversarial_2021, sun_certifiably_2023, xue_mis-spoke_2022}, or target multi-agent systems with few agents so that the selection of which messages to alter is not a consideration \cite{ma_grey-box_2023}. Likewise, when to attack has only been explored against single-agent systems \cite{lin_tactics_2017, sun_stealthy_2020, kos_delving_2017, qiaoben_strategically-timed_2021, zheng_vulnerability_2021, praveen_kumar_critical_2021}. The attack tempo, which determines when an attack occurs, can have a major impact on its effectiveness \cite{lin_tactics_2017, kos_delving_2017, sun_stealthy_2020, qiaoben_strategically-timed_2021, zheng_vulnerability_2021, praveen_kumar_critical_2021}. An effective attack tempo against multi-agent communications is challenging due to the lower success rate of the attack. Thus, the tempo of communication attacks should consider both the potential success and the impact of an attack. Our work proposes methods for efficiently identifying which messages to perturb, which agent to target, and when to perturb to maximise attack effectiveness, by exploiting white-box knowledge of the target system and extending existing tempo methods to the multi-agent domain.

The diversity of valid solutions to Reinforcement Learning (RL) problems introduces additional challenges in attacking Multi-Agent RL (MARL) systems. A successful attack should impact the target system, which white-box AML attacks against non-RL systems often achieve with gradient-based perturbations that alter the output of the target \cite{goodfellow_explaining_2015, madry_towards_2018, tu_adversarial_2021}. However, just changing the action taken by an agent may not be sufficient to impact the system because alternative actions may lead to similar outcomes. To overcome this problem, many approaches train a neural network using deep RL that can identify the actions with the highest impact that are then induced by gradient-based observation perturbations \cite{lin_robustness_2020, sun_who_2022, wan_exploring_2022, qiaoben_understanding_2024} or the network can directly perturb observations \cite{russo_towards_2021, garcia_learning_2020, pattanaik_robust_2018, zhang_robust_2020-1} and messages \cite{sun_certifiably_2023, xue_mis-spoke_2022} to cause high-impact actions. The limitation of these attacks is their reliance on high-compute resources and narrow application to a specific target. Instead, we extend gradient-based perturbation crafting techniques with new loss functions to find successful attacks that cause a meaningful impact on system performance, without costly techniques such as deep RL.

Our approach proposes a model of a single-victim communication attack, a specific attack scenario in which the adversary alters the messages received by a single agent in the system \cite{sun_certifiably_2023}. This model allows us to identify key unaddressed aspects of the communication attack and expand existing solutions. 
We propose two new loss functions, namely, \textit{weighted-loss} and \textit{maximum-loss}. Weighted-loss attempts to maximise attack effectiveness by optimising both the impact of an attack and the probability of success. The maximum loss sacrifices the probability of success in return for a greater impact on the system. 
The problems of selecting where to attack, who to attack, and when to attack have not been addressed for attacks against MARL in any previous literature, so we propose a method that identifies which messages, agents, and timesteps an adversary should attack to maximise attack effectiveness using an approach inspired by Jacobian-based saliency methods \cite{papernot_limitations_2016}. 
The standard Jacobian Saliency Map Attack (JSMA) uses an output-targeted loss function to compute the Jacobian, which is used to target the most influential elements of the input. Our method  computes the Jacobian with respect to our proposed loss functions and uses the resulting gradient magnitudes to select the top-k most influential messages. We also use the gradient magnitudes of the messages to identify which agent and which timesteps are most susceptible to attack.
Our experiments demonstrate the improved impact of our methods, particularly in more complex environments and against more robust communication methods.
In summary, our contributions are three-fold:
\begin{itemize}
    \item A framework for single-victim communication attacks that exposes overlooked weaknesses in inter-agent communication
    \item Two loss functions that enhance the effectiveness of perturbations against MARL
    \item A gradient-based approach that identifies the most vulnerable messages, agents, and timesteps that are the weakest links in MARL systems
\end{itemize}

\section{Background and Related Work}
\label{sec:related_work}

We provide an initial background to MARL and multi-agent communication and then discuss existing communication perturbation attacks, perturbation crafting methods, and tempo functions used to determine the tempo of attacks of deep RL systems. 

MARL extends deep RL to multi-agent problems and a unique aspect of multi-agent problems is communication, which is necessary for coordination and information sharing between agents. A common approach to communication allows agents to share their observations \cite{kilinc_multi-agent_2018,gupta_hammer_2023,kong_revisiting_2017}. However, this may be impractical in real-world scenarios due to bandwidth or Size Weight And Power (SWAP) constraints. Instead, more efficient communication protocols can be learnt using MARL (MARL-Comms) \cite{zhu_survey_2024}. The key properties of these algorithms for an attacker are the frequency of communication and the structure of the messages sent between agents. Reinforced Inter-Agent Learning (RIAL) \cite{foerster_learning_2016-1} is a MARL-Comms algorithm that treats communication as an auxiliary reinforcement learning task and uses Deep Q-Networks (DQN) \cite{mnih_playing_2013} to simultaneously learn action and message policies. This communication occurs once between timesteps, and the message is selected from a discrete set of possible messages. This contrasts with an approach such as CommNet \cite{sukhbaatar_learning_2016}, which features multiple rounds of communication during a timestep with a continuous set of messages.

There have been a variety of AML attacks targeting MARL \cite{standen_adversarial_2025}, but relatively few communication perturbation attacks, which occur when an adversary intercepts and perturbs the messages sent between agents. We consider communication perturbation attacks separate to malicious communication attacks as identified by Standen et al. \cite{standen_adversarial_2025}, in which an adversary can inject new malicious messages into the system.
Previous communication perturbations attacks targeted both MARL \cite{sun_certifiably_2023, xue_mis-spoke_2022, ma_grey-box_2023} and non-RL multi-agent \cite{tu_adversarial_2021} systems. The messages the adversary intercepts is a unique aspect of these attacks and there have been two approaches to selecting these messages, namely: \textit{broadcast} and \textit{single victim}. Broadcast attacks perturb a subset of messages that are broadcast to all agents in the system  \cite{tu_adversarial_2021, xue_mis-spoke_2022, ma_grey-box_2023}. Single-victim attacks perturb a subset of messages received by a single agent in the system \cite{sun_certifiably_2023}. However, none of these works address which messages an adversary should perturb to maximise the effectiveness of an attack. Instead, they either preselect messages \cite{tu_adversarial_2021, sun_certifiably_2023, xue_mis-spoke_2022} without consideration of attack effectiveness, or attacked small multi-agent systems where there is only a single message to perturb \cite{ma_grey-box_2023}.

Communication perturbation attacks rely on cleverly crafted perturbations that affect behaviour of a system. An adversary with \textit{white-box} knowledge of the victim may use gradient-based methods to craft these perturbations. Single-step gradient methods, such as Fast Gradient Method (FGM) \cite{goodfellow_explaining_2015}, uses the gradient of the input with respect to a particular loss function, and assumes a linear response from the network to find an effective perturbation. Another gradient-method Projected Gradient Descent (PGD) \cite{madry_towards_2018}, iteratively applies FGM, allowing perturbations to correct for the non-linear response of the network. The default loss function used in these attacks is called the untargeted loss and aims to minimise the probability that the agent will output the same action as it would with unperturbed input. However, for RL systems, these untargeted attacks may cause the victim to select an alternate action with similar outcomes, thus the attack causes minimal impact against the system \cite{lin_robustness_2020}.
Instead, two-stage attacks first learn an adversarial policy which minimises the system reward and then uses a gradient method to cause the victim to output the actions selected by the adversarial policy \cite{lin_robustness_2020, sun_who_2022, wan_exploring_2022, qiaoben_understanding_2024}. 
Extending on this idea, RL-based attacks use deep RL to directly learn effective perturbations \cite{russo_towards_2021, garcia_learning_2020, sun_certifiably_2023, xue_mis-spoke_2022, pattanaik_robust_2018, zhang_robust_2020-1}, forgoing the need of gradient-based perturbation crafting methods. 
However, RL can require significantly more time and computation than the original untargeted perturbation and the learnt functions can only target a specific environment or system.

Adversarial attacks against MARL need to consider when to attack. Selecting an effective tempo allows an adversary to minimise the number of attacked timesteps without compromising attack effectiveness. Current tempo methods all target single-agent systems, and no published work, to the best of our knowledge, has considered the tempo of attacks against MARL.
\textit{Fixed tempos} do not consider the attack effectiveness and include tempos that attack every timestep \cite{huang_adversarial_2017} or attack a contiguous set of timesteps \cite{lin_tactics_2017}. To improve attack effectiveness, \textit{counterfactual tempos} simulate an attack \cite{sun_stealthy_2020}, and \textit{learnt tempos} train a deep RL agent to learn when to attack \cite{sun_stealthy_2020}. However, these methods can be costly due to the simulation and training time respectively. \textit{Threshold tempos} measure certain properties of an agent's logits, and attack when that metric exceeds a hyperparameter threshold \cite{lin_tactics_2017, kos_delving_2017, qiaoben_strategically-timed_2021, zheng_vulnerability_2021, praveen_kumar_critical_2021}. Criticality-Based Timing Selection (CBTS) \cite{zheng_vulnerability_2021} uses the difference between the first and second highest logits, \(\max(Q)-\max_2(Q)\).
Max-Min Ratio (MMR) \cite{praveen_kumar_critical_2021} uses the ratio between the maximum and minimum logits, \(\displaystyle \frac{\max(Q)}{1+\min(Q)}\).
Maximum Logit (ML) \cite{praveen_kumar_critical_2021} uses the maximum logit, \(\max(Q)\).
Negative Skew (NS) \cite{praveen_kumar_critical_2021} uses the negative skew in the logit distribution, \(\frac{3 \times (\mean(Q) - \median(Q))}{ \sigma(Q)}\).
Variance of Logits (VL) \cite{praveen_kumar_critical_2021} uses the variance of the logits, \(\sigma(Q)\).
Strategically-Timed (ST) \cite{lin_tactics_2017} uses the difference between largest and smallest logits, \(\max(Q) - \min(Q)\). A drawback of threshold tempos is the requirement to specify the attack tempo as a hyperparameter. 

In summary, while there has been some work looking at communication perturbation attacks \cite{sun_certifiably_2023, xue_mis-spoke_2022, ma_grey-box_2023, tu_adversarial_2021}, these attacks do not address the key question of which messages an adversary should perturb to maximise attack effectiveness. The crafting of communication perturbations against MARL has heavily relied on deep RL-based methods which are costly to train. The question of when to attack has also not been addressed in previous communication perturbation attacks. In this work, we address these gaps by proposing a message selection method that identifies which messages should be perturbed to maximise the effectiveness of the attack, proposing new loss functions to improve the effectiveness of gradient-based perturbation crafting methods, and exploring victim selection methods by extending existing tempo functions and proposing a new tempo function.

\section{Problem Definition}

To explore the worst-case vulnerability of multi-agent communication, we assume a strong attacker capability. We assume the attacker has full white-box knowledge of the victim system and can intercept and alter inter-agent messages.

Attackers targeting communication in multi-agent systems have two goals, namely, maximise the effectiveness of the attack and minimise its detectability. Attack effectiveness is determined by \textit{attack success} and \textit{attack impact}. Attack success is the ability of an attack to alter the output of the victim which for MARL is the action selected by the victim. Attack impact is the effect of a successful attack on the performance of the system. 

The detectability of an attack against multi-agent communications depends on the magnitude of change, the tempo of the attack, and the number of changed messages per attacked timestep. We focus on the single-victim scenario, where the attacker only changes a subset of messages received by a single agent in the system \cite{sun_certifiably_2023}.

To model this problem, we use a variant of the Adversarial Partially Observable Stochastic Game (APOSG) \cite{standen_adversarial_2025} that we call the Single-Victim Communication Perturbation APOSG (SVCP-APOSG). The SVCP-APOSG is defined by the 14-tuple $\{I, S, \mathcal{M}, \hat{A}, T, \\\hat{\Omega}, \hat{O}, \hat{R}, v^m, \Theta^m, \Delta^m, \Sigma_a, \Sigma_m, k\}$, which we can separate into environmental and adversarial elements.

The eight environmental elements in the SVCP-APOSG are $\{I, S, \\ \mathcal{M}, \hat{A}, T, \hat{\Omega}, \hat{O}, \hat{R}\}$ and determine the training environment of the target system.
$I$ is the set of agents. 
$S$ is the set of states. 
$\mathcal{M}$ is the shared message space.
$\hat{A}$ is the joint action sets. At each timestep, agents select both an environmental action, that affects the next state of the environment, and a message action, which determines the information communicated to the other agents in the system.
$T$ is the state transition function that determines the probability of transitioning to a next state and is only dependent on current state and the environmental actions selected by the agents.
$\hat{\Omega}$ is the joint observation space. Each observation contains environmental information and the messages received by an agent.
$\hat{O}$ is the joint observation probability function.
$\hat{R}$ is the set of reward functions for each agent. 

The six adversarial elements of the SVCP-APOSG are $\{v^m, \Theta^m, \Delta^m, \\ \Sigma_a, \Sigma_m, k\}$ and determine the behaviour of a particular attack.
$v^m:\mathcal{M}\rightarrow\mathcal{M}$ is the message-permutation function, which calculates the perturbed messages received by a victim agent.
$\Theta^m: S \rightarrow[0,1]$ is the message attack tempo function that determines the probability that a timestep will be attacked.
$\Delta^m: \mathbb{R}^+$ is the message attack magnitude, which is the maximum $L^2$ distance between the perturbed message and the original message.
$\Sigma_a: S \rightarrow I$ is the victim selection function, which determines the victim of an attack.
$\Sigma_m: S \rightarrow \{I\}^k$ is the message selection function, which determines which of the received messages are perturbed.
$k: \mathbb{Z}^+$ is the number of messages in the message observation that are perturbed.
For each of the adversarial aspects, the adversary wants to maximise the attack effectiveness. Additionally, the adversary wants to minimise attack detectability by minimising $\Delta^m$, $k$, and the number of perturbed timesteps as determined by $\Theta^m$.

We use the joint observation $\hat{o}$ to refer to the set of observations received by all agents in the system and $o_i$ to refer to the observation of a agent $i$ within that joint set. This observation is composed of the environment observation $o^e_i$ and the messages received by the agent $[o^m_1, ..., o^m_{|I|}]$

We only consider agents trained with Q-learning approaches and assume that the agents are well trained and accurately predict the Q-value for each environmental action that they can take. The Q-value function learnt by an agent is $Q(a_{t}|o_{t}) = r_t + \gamma \mathbb{E} \max_{a' \in A}(Q(a'|o_{t+1}))$, where $a_{t,i}$ is the environmental action of an agent at timestep $t$, $A$ is the set of environmental actions of the agent, and, $o_{t,i}$ and $o_{t+1, i}$ are the observations for timesteps $t$ and $t+1$ respectively. Thus, the logits of an agent are that agent's estimate of the Q-value for each action. For brevity, we will denote the full distribution of logits using $Q$ and the Q-value of a specific action $a$ as $Q(a)$. We often refer to the Q-difference which is the difference between the maximum predicted Q-value and the Q-value of a specific action $a$, such that $Q_{\text{diff}}(a) = \max(Q)-Q(a)$. 

\section{Attack Design}

In this section, we describe the design of our AML attack. 
We propose two loss functions, \textit{weighted loss} and \textit{maximum loss}, which aim to improve the effectiveness of gradient-based perturbation crafting functions $v^m$. 
We address the three key components mentioned above, namely, where-to-attack, who-to-attack, and when-to-attack, represented as $\Sigma_m$ $\Sigma_a$, and $\Theta^m$ in the SVCP-APOSG respectively, by proposing a method that uses a Jacobian-proxy value to select which messages, victims, and timesteps will maximise attack effectiveness. 

\subsection{Loss functions}

To maximise attack effectiveness, we propose two novel loss functions that can be combined with gradient-based perturbation crafting methods such as FGM \cite{goodfellow_explaining_2015} and PGD \cite{madry_towards_2018} to improve attack effectiveness over the untargeted loss function. 

The untargeted loss function, used in previous attacks \cite{huang_adversarial_2017, kos_delving_2017}, uses the cross-entropy loss between an agent's logits and the argmax of the logits to minimise the probability that the agent will output the same action eq. (\ref{eq: untargeted}). However, this loss function may cause the agent to select an alternative action with a similar outcome to the original action. This is because the untargeted loss considers any alternative action as good as any other action. Against a well-trained agent that knows multiple paths to success, this may cause little to no impact on the system.

\begin{equation}
    L_u(o_i) = - \log(\frac{e^{\max(Q)}}{\sum_{a \in A}e^{Q(a)}})
    \label{eq: untargeted}
\end{equation}

To improve the impact of the attack, we propose a loss function to encourage the agent to select the action that was originally considered to be the worst action by maximising the probability of that action eq.(\ref{eq: maximum}), which we call the maximum loss, $L_m$. If successful, perturbations caused by $L_m$ will cause the maximum impact against a well-trained agent. However, the improved attack impact likely comes with a reduced chance of attack success because actions with higher $Q_{\text{diff}}$ values may be harder to induce. 

\begin{equation}
    L_{m}(o_i) = \log(\frac{e^{\max(Q_{\text{diff}})}}{\sum_{a \in A}e^{Q(a)}})
    \label{eq: maximum}
\end{equation}

To balance attack impact and attack success, we propose an alternative loss function, that we call weighted loss, $L_w$ (eq.\ref{eq: weighted}), which maximises the probability of each action relative to its $Q_{\text{diff}}$ by using a mean weighted cross-entropy loss across all actions.

\begin{equation}
    L_{w}(o_i) = \frac{1}{\sum_{a \in A} Q_{\text{diff}}(a)} \sum_{a \in A} Q_{\text{diff}}(a) \log(\frac{e^{Q(a)}}{\sum_{b \in A}e^{Q(b)}})
    \label{eq: weighted}
\end{equation}

\subsection{Jacobian-Magnitude for Message selection, Victim selection and Tempo functions}
To select the weakest links, we need methods capable of measuring the vulnerability of each element. Measuring timestep vulnerability may be done using the threshold tempo metrics, and these metrics could also be used to measure the vulnerability of an agent, allowing for victim selection. However, these techniques do not allow us to identify which messages should be attacked and so we propose a method inspired by JSMA \cite{papernot_limitations_2016}, that uses the Jacobian of the messages received by an agent as a proxy for the potential attack effectiveness of a communication perturbation attack. This proxy can then be used to determine which message, victim, and timesteps to attack.
Previous tempo methods all focus on measuring the attack impact to determine which timesteps to attack, whereas our Jacobian-proxy method measures the attack success for the untargeted loss function or attack effectiveness for the maximum and weighted loss functions.

We denote the Jacobian as $J(o_i)=\nabla L(o_i)$, where $L$ can be any loss function including those in eq. \ref{eq: untargeted}, eq. \ref{eq: maximum} and eq. \ref{eq: weighted}. The Jacobian can be separated into components corresponding to elements of the observation $o_i$ received by agent $i$, such that $J(o_i)=[J(o^e_i),J(o^m_{i,1}),...,J(o^m_{i,|I|})]$. For brevity, we define the element of the Jacobian corresponding to the message received by agent $i$ from agent $j$ $J(o^m_{i,j})$ as $J^m_{i,j}$. We then take the element-wise summation of the absolute values of each message Jacobian as the proxy for attack effectiveness against that message such that our proxy $P:\Omega\times I\rightarrow\mathbb{R}^+$ is
\begin{equation}
P(o_i,j) = \sum_{k}|J^m(o_i,j)_k|
\label{eq: jacobian_proxy}
\end{equation}

Our proposed message selection function, that we call ranked message selection, ranks each message received by its Jacobian magnitude and selects the top-$k$ messages. Our message selection function for the victim agent $i$ in
\begin{equation}
    \Sigma_m(o_i) = \topk_k(\{P(o_i,j); j \in I\})
    \label{eq: ranked_msg}
\end{equation}
where the function $\topk_k$ returns the indices of the highest $k$ values. We use the observation function $O$ to get the observation $o_i$ for agent $i$ from the state $s$.

Our victim selection function selects the agent with the largest total Jacobian magnitude of the top $k$ messages as shown in 
\begin{equation}
    \Sigma_a(\hat{o}) = \arg\max_{i \in I}[\sum_{j \in \Sigma_m(o_i)} P(o_i,j)]
    \label{eq: victim_selection}
\end{equation}
such that we find the total Jacobian-proxy value of the messages selected by $\Sigma_m(o_i)$ for each agents $i$, and select the victim to have the largest total Jacobian-proxy value.

Our tempo function selects the timesteps where the magnitude of the Jacobian of the top $k$ messages received from the selected victim $i$ exceeds the hyperparameter threshold $\phi$ as shown in
\begin{equation}
    \Theta(o_i) = \mathbf{1}\{[\sum_{j \in \Sigma_m(o_i)} P(o_i,j)] > \phi\}
    \label{eq: jacobian_tempo}
\end{equation}

\section{Experimental Analysis}
\label{sec:experiments}
For our experiments, we use five environments, namely, a simple grid world navigation game \cite{singh_learning_2018}, two variants of the PredatorPrey game \cite{xue_mis-spoke_2022} and two variants of the TrafficJunction game \cite{xue_mis-spoke_2022, sukhbaatar_learning_2016}.
The multi-agent system controls three agents in the navigation and PredatorPrey environments, a maximum of five agents in the small TrafficJunction environment, and a maximum of 20 in the large TrafficJunction environment.

In our navigation game, the agents need to locate and move to a particular location in the world. Each agent has a limited view around itself and may move in any of the four cardinal directions. The reward function is $R = 0.05\frac{n_{goal}^2 + n_{goal} - n}{n} $, where $n=3$ is the number of agents in the environment and $n_{goal}$ is the number of agents at the landmark.

Our first variant of PredatorPrey, we call orthogonal PredatorPrey, also uses a grid world with the same observation and movements as the navigation game. The predators, controlled by the target system must cooperate to catch six prey that move randomly around the environment.
The second variant of PredatorPrey, we call diagonal PredatorPrey, uses the same setup as the orthogonal PredatorPrey environment, except with a larger action space that allows additional diagonal movements by the predators and prey, and the locations of each agent are no longer fixed to a grid. 
Both PredatorPrey environments use the same reward function of $R = 10*c_s - 0.5 c_f$, where $c_s$ is a successful catch of a prey, and $c_f$ is the number of agents that performed the catch action but did not catch prey.

In the TrafficJunction environments, the agents follow routes that intersect, and they must coordinate their movements to prevent collisions. The agents only have two actions, namely, stop and go, and may only observe agents within a limited range. The small environment only has one intersection, and the large environment has four.
Both TrafficJunction environments use the reward function $R = -10c-0.01t$, where $c$ is the number of collisions, and $t$ is the total lifetime of the agents in the system. A collision occurs when two agents share a tile.

We share our code, data, and additional experimental details at <GitHub link redacted>. For review: we have attached our code and data to our submission and intend on making that code and data available on GitHub after publication.

We evaluate our attacks against two communication methods: full observation sharing (OBS) and RIAL \cite{foerster_learning_2016-1}. The structure of the messages for full observation sharing is dependent on the environment and is a standard multi-agent communication method \cite{kilinc_multi-agent_2018, gupta_hammer_2023, kong_revisiting_2017}. For RIAL, we use a vocabulary size of four and message length of two. The specific messages in our set are $[(1.0, 1.0),(1.0,-1.0),(-1.0,1.0)],(-1.0,-1.0)]$. Our message space is $\mathcal{M} = \mathbb{R}^l$, where $l=2$ for RIAL, and $l=|\Omega|$ for observation sharing. This continuous message space allows us add $L^2$ perturbations to both communication methods.

We conduct two experiments to empirically evaluate the effectiveness of our methods. First, we compare our attacks with other baseline attacks that do not use any of our methods. The baseline attacks use the tempo methods CBTS \cite{zheng_vulnerability_2021}, MMR \cite{praveen_kumar_critical_2021}, ML \cite{praveen_kumar_critical_2021}, NS \cite{praveen_kumar_critical_2021}, VL \cite{praveen_kumar_critical_2021}, and ST \cite{lin_tactics_2017} as both tempo and victim selection functions and use them in combination with random message selection and PGD \cite{madry_towards_2018} with the untargeted loss function for perturbation crafting. We extend the tempo methods to victim selection methods by selecting the victim with the highest metric. In the TrafficJunction environment, the action space size makes CBTS, ST, and VL equivalent, so we only present the results for ST. This does not apply to RIAL because the action space includes both the environmental and message selection actions.
Secondly, we perform an ablation study on the loss function and message selection methods by considering attacks that use maximum, weighted, or untargeted loss, with random and ranked message selection. 
For each of these experiments, we used an attack magnitude $\Delta^m = 1$, which we identified from a grid search of attack magnitudes. To control the attack frequency for a fair comparison between attacks, we use a binning method, in which we use three bins, namely, 0.25, 0.5, and 0.75, with widths of 0.125. The attack frequency is the proportion of attacked steps to the length of an episode. To collect data for each of these bins we experimentally selected thresholds for the tempo techniques by collecting data from unattacked episodes. All attacks use PGD \cite{madry_towards_2018} with their respective loss functions to craft the perturbations with 20 iterations and a step size of 0.1.

When performing our experiments, to measure attack effectiveness, we measure the total reward of an episode. We also use the task metric, which is an idea we derive from the T-function presented by Sun et al. \cite{sun_stealthy_2020} to measure a specific element of the task being performed by the system. For navigation, the task metric is the proportion of agents that reach the goal by the end of an episode. For PredatorPrey, the task metric is the proportion of caught prey. For TrafficJunction, it is the number of collisions. We present the clean reward and task metric in our results which is the value achieved by the unattacked system.
Finally, we measure attack success rate using the proportion of attacked steps where the action of the victim changed $\Delta a = \frac{1}{|T'|} \sum_{t \in T'}\mathbb{1}(a_t\neq a'_t)$, where $T'$ is the set of attacked timesteps, $a_t$ is the original action the agent would have performed at timestep $t$ and $a'_t$ is the adversarially-induced action at timestep $t$.

In the presentation of our results, we use abbreviations due to space constraints.
We abbreviate the navigation environment to Nav, the orthogonal and diagonal PredatorPrey environments to PP-O and PP-D respectively, and the small and large TrafficJunction environments to TJ-S and TJ-L respectively.
We denote our attacks as J-W and J-M, which indicate the use of weighted and maximum loss functions respectively along with our Jacobian tempo, victim, and messages selection methods.

In presenting the reward and t-metric results of our experiments, we use Tukey's honest significance test with $\alpha=0.05$ and bold the results that have the greatest impact on the system and the results that are not significantly different to the results with the greatest impact unless the result is not significantly different to the clean result.

\section{Results}

We present the impact of our attacks and the baseline attacks on the reward, the success rate of the attack, and the impact on the task metric against the observation sharing system in Tables \ref{tab:ObservationSharing_overview_reward}, \ref{tab:ObservationSharing_overview_action_changed} and \ref{tab:ObservationSharing_overview_t_metric} respectively.
In the navigation environment, ML, and NS are effective at moderate attack rates but lose their effectiveness at higher attack rates. CBTS and our J-M attack achieve a significant effect at high attack rates.
The trends in the PredatorPrey environments are simpler with all attacks increasing or maintaining their effectiveness as the attack rate increases. 
In the orthogonal PredatorPrey environment, our J-M and J-W attacks are the most effective across all attack rates and are joined by CBTS at low and moderate attack rates, and NS at high attack rates.
For the diagonal PredatorPrey, J-W is the most effective for all attack rates but only offers a significant improvement over baseline attacks at high attack rates.
For the TrafficJunction environments, J-W and J-M are the most effective attacks except for high attack rates against the large environment, in which ST is the most effective and our Jacobian attacks lose their effectiveness.

We present our results against RIAL in Tables \ref{tab:RIAL_overview_reward}, \ref{tab:RIAL_overview_action_changed}, and \ref{tab:RIAL_overview_t_metric} which show the reward, attack success rate, and task metric respectively.
We see similar trends in the RIAL results as we did in the observation sharing results, with some of the attacks only achieving a significant impact at higher attack rates, while others lose their effectiveness as the attack rate increases. 
In the navigation environment, ML and NS are most effective at low attack rates, our J-M and J-W attacks are most effective at moderate and high attack rates and are joined by MMR at moderate attack rates and CBTS, VL, and ST at high attack rates.
In the orthogonal PredatorPrey environment, only ML at high attack rate achieved a significant impact. 
J-M achieved a significant impact against the diagonal PredatorPrey at moderate and high attack rates. 
Against the small TrafficJunction environment, ML and MMR were the most effective at low and moderate attack rates, and J-M, J-W, VL, and ST were the most effective at high attack rates. In the large TrafficJunction environment, we see a similar trend with ML being the most effective at low and moderate attack rates, and MMR being the most effective at moderate and high attack rates. However, the impacts across the PredatorPrey and TrafficJunction environments are much smaller than those that were induced against the observation sharing system and none of the attacks achieve a significant impact on the task metric of the PredatorPrey and small TrafficJunction environments. Instead, the attacks delay the success of the system, reducing the reward.
The robustness of RIAL to attack is also demonstrated by the low attack success rates across all environments shown in Table \ref{tab:RIAL_overview_action_changed}.

Another factor in the robustness of RIAL are local optima, which cause the adversarially induced actions to improve the performance of the system. We observe this phenomenon in the observation sharing system in the navigation environment, and in all RIAL systems. This highlights an issue with our assumption that the system is well trained because the benign action of a well-trained system should always be the best action, thus a malicious action should never be able to increase the performance of the system.

\begin{table}
	\centering
	\caption{Reward of the observation sharing systems. Clean reward shown in brackets next to the environment name.}
	\label{tab:ObservationSharing_overview_reward}
	\begin{tabular}{lccccccccc}\toprule
		  & \multicolumn{3}{l}{{Nav (\textit{1.56})}} & \multicolumn{3}{l}{{PP-O (\textit{42.4})}} & \multicolumn{3}{l}{{PP-D (\textit{35.6})}}\\\cmidrule(lr){2-4} \cmidrule(lr){5-7} \cmidrule(lr){8-10}
		Attack name & 0.25 & 0.5 & 0.75 & 0.25 & 0.5 & 0.75 & 0.25 & 0.5 & 0.75 \\ \midrule
		J-M (Ours) & 1.98 & 1.39 & \textbf{0.48} & \textbf{38.8} & \textbf{34.6} & \textbf{31.4} & 28.6 & 24.0 & 19.4\\
		J-W (Ours) & 1.97 & 1.46 & 0.7 & \textbf{39.6} & \textbf{35.7} & \textbf{31.5} & \textbf{24.0} & \textbf{21.7} & \textbf{17.1}\\
		CBTS \cite{zheng_vulnerability_2021} & 2.01 & 1.49 & \textbf{0.54} & \textbf{38.3} & \textbf{35.6} & 33.2 & \textbf{25.9} & \textbf{22.5} & 19.9\\
		MMR \cite{praveen_kumar_critical_2021} & 1.56 & 1.16 & 0.81 & 40.5 & 37.5 & 36.9 & 27.0 & \textbf{22.8} & 20.2\\
		ML \cite{praveen_kumar_critical_2021} & 1.62 & \textbf{0.98} & 1.26 & 41.7 & 38.1 & 33.8 & \textbf{25.7} & 23.6 & 20.8\\
		NS \cite{praveen_kumar_critical_2021} & 1.7 & \textbf{0.83} & 1.19 & 40.4 & 37.7 & \textbf{32.7} & 31.8 & 27.1 & 24.0\\
		VL \cite{praveen_kumar_critical_2021} & 2.02 & 1.55 & 0.71 & 40.6 & 38.5 & 36.7 & 26.2 & 24.0 & 19.8\\
		ST \cite{lin_tactics_2017} & 2.04 & 1.52 & 0.72 & 40.7 & 37.8 & 36.2 & 26.2 & 23.5 & 20.3\\
		\bottomrule
	\end{tabular}
	\begin{tabular}{lcccccc}\toprule
		  & \multicolumn{3}{l}{{TJ-S (\textit{-1.33})}} & \multicolumn{3}{l}{{TJ-L (\textit{-28.2})}}\\\cmidrule(lr){2-4} \cmidrule(lr){5-7}
		Attack name & 0.25 & 0.5 & 0.75 & 0.25 & 0.5 & 0.75 \\ \midrule
		J-M (Ours) & \textbf{-3.24} & \textbf{-9.82} & \textbf{-18.3} & \textbf{-152.2} & \textbf{-218.4} & -177.6\\
		J-W (Ours) & \textbf{-3.62} & \textbf{-8.77} & \textbf{-18.09} & \textbf{-169.8} & \textbf{-210.6} & -190.4\\
		MMR \cite{praveen_kumar_critical_2021} & -1.42 & -1.54 & -1.48 & -29.1 & -29.4 & -29.2\\
		ML \cite{praveen_kumar_critical_2021} & -2.76 & -4.15 & -3.65 & -68.8 & -58.0 & -43.6\\
		NS \cite{praveen_kumar_critical_2021} & -1.61 & -1.59 & -1.59 & -28.5 & -29.5 & -30.8\\
		ST \cite{lin_tactics_2017} & -2.78 & -8.03 & -16.64 & -110.5 & -188.7 & \textbf{-261.1}\\
		\bottomrule
	\end{tabular}
\end{table}
\begin{table}
	\centering
	\caption{{Rate of successful action changes against the observation sharing systems.}}
	\label{tab:ObservationSharing_overview_action_changed}
	\begin{tabular}{lccccccccc}\toprule
		  & \multicolumn{3}{l}{Nav} & \multicolumn{3}{l}{PP-O} & \multicolumn{3}{l}{PP-D}\\\cmidrule(lr){2-4} \cmidrule(lr){5-7} \cmidrule(lr){8-10}
		Attack name & 0.25 & 0.5 & 0.75 & 0.25 & 0.5 & 0.75 & 0.25 & 0.5 & 0.75 \\ \midrule
		J-M (Ours) & 0.31 & 0.41 & 0.49 & 0.68 & 0.61 & 0.56 & 0.8 & 0.82 & 0.84\\
		J-W (Ours) & 0.44 & 0.5 & 0.62 & 0.68 & 0.6 & 0.57 & 0.88 & 0.89 & 0.91\\
		CBTS \cite{zheng_vulnerability_2021} & 0.27 & 0.43 & 0.55 & 0.41 & 0.42 & 0.45 & 0.81 & 0.91 & 0.94\\
		MMR \cite{praveen_kumar_critical_2021} & 0.38 & 0.51 & 0.62 & 0.49 & 0.55 & 0.59 & 0.83 & 0.91 & 0.93\\
		ML \cite{praveen_kumar_critical_2021} & 0.52 & 0.62 & 0.66 & 0.6 & 0.64 & 0.64 & 0.84 & 0.92 & 0.95\\
		NS \cite{praveen_kumar_critical_2021} & 0.74 & 0.78 & 0.79 & 0.77 & 0.71 & 0.7 & 1.0 & 0.99 & 0.99\\
		VL \cite{praveen_kumar_critical_2021} & 0.39 & 0.5 & 0.61 & 0.52 & 0.56 & 0.6 & 0.82 & 0.9 & 0.93\\
		ST \cite{lin_tactics_2017} & 0.36 & 0.5 & 0.59 & 0.48 & 0.54 & 0.57 & 0.82 & 0.9 & 0.93\\
		\bottomrule
	\end{tabular}
	\begin{tabular}{lcccccc}\toprule
		  & \multicolumn{3}{l}{TJ-S} & \multicolumn{3}{l}{TJ-L}\\\cmidrule(lr){2-4} \cmidrule(lr){5-7}
		Attack name & 0.25 & 0.5 & 0.75 & 0.25 & 0.5 & 0.75 \\ \midrule
		J-M (Ours) & 0.39 & 0.55 & 0.63 & 0.99 & 0.99 & 0.99\\
		J-W (Ours) & 0.39 & 0.54 & 0.63 & 0.99 & 0.99 & 0.99\\
		MMR \cite{praveen_kumar_critical_2021} & 1.0 & 1.0 & 1.0 & 1.0 & 1.0 & 1.0\\
		ML \cite{praveen_kumar_critical_2021} & 0.94 & 0.91 & 0.92 & 1.0 & 1.0 & 1.0\\
		NS \cite{praveen_kumar_critical_2021} & 1.0 & 1.0 & 1.0 & 1.0 & 1.0 & 1.0\\
		ST \cite{lin_tactics_2017} & 0.23 & 0.38 & 0.54 & 0.96 & 0.96 & 0.96\\
		\bottomrule
	\end{tabular}
\end{table}
\begin{table}
	\centering
	\caption{{Task metric of the observation sharing systems. Clean task metric shown in brackets next to the environment name.}}
	\label{tab:ObservationSharing_overview_t_metric}
	\begin{tabular}{lccccccccc}\toprule
		  & \multicolumn{3}{l}{{Nav (\textit{2.7})}} & \multicolumn{3}{l}{{PP-O (\textit{4.68})}} & \multicolumn{3}{l}{{PP-D (\textit{3.96})}}\\\cmidrule(lr){2-4} \cmidrule(lr){5-7} \cmidrule(lr){8-10}
		Attack name & 0.25 & 0.5 & 0.75 & 0.25 & 0.5 & 0.75 & 0.25 & 0.5 & 0.75 \\ \midrule
		J-M (Ours) & 2.88 & 2.76 & \textbf{2.27} & 4.55 & 4.38 & 4.24 & 3.46 & 3.23 & 2.94\\
		J-W (Ours) & 2.87 & 2.83 & 2.5 & \textbf{4.47} & \textbf{4.24} & \textbf{4.02} & \textbf{2.87} & \textbf{2.74} & \textbf{2.39}\\
		CBTS \cite{zheng_vulnerability_2021} & 2.88 & 2.88 & 2.46 & \textbf{4.37} & \textbf{4.3} & 4.25 & \textbf{3.0} & \textbf{2.65} & \textbf{2.47}\\
		MMR \cite{praveen_kumar_critical_2021} & 2.73 & 2.72 & 2.59 & 4.57 & 4.39 & 4.39 & 3.22 & \textbf{2.79} & 2.6\\
		ML \cite{praveen_kumar_critical_2021} & 2.81 & 2.7 & 2.84 & 4.64 & 4.51 & 4.34 & \textbf{2.95} & \textbf{2.81} & 2.56\\
		NS \cite{praveen_kumar_critical_2021} & 2.81 & \textbf{2.47} & 2.53 & 4.63 & 4.54 & 4.3 & 3.6 & 3.22 & 3.0\\
		VL \cite{praveen_kumar_critical_2021} & 2.88 & 2.88 & 2.55 & 4.54 & 4.46 & 4.41 & 3.11 & 2.89 & 2.54\\
		ST \cite{lin_tactics_2017} & 2.89 & 2.87 & 2.62 & 4.55 & 4.42 & 4.39 & \textbf{3.07} & \textbf{2.82} & 2.58\\
		\bottomrule
	\end{tabular}
	\begin{tabular}{lcccccc}\toprule
		  & \multicolumn{3}{l}{{TJ-S (\textit{0.01})}} & \multicolumn{3}{l}{{TJ-L (\textit{0.14})}}\\\cmidrule(lr){2-4} \cmidrule(lr){5-7}
		Attack name & 0.25 & 0.5 & 0.75 & 0.25 & 0.5 & 0.75 \\ \midrule
		J-M (Ours) & \textbf{0.2} & \textbf{0.83} & \textbf{1.64} & \textbf{11.94} & \textbf{18.03} & 14.06\\
		J-W (Ours) & \textbf{0.24} & \textbf{0.73} & \textbf{1.63} & \textbf{13.55} & \textbf{17.29} & 15.29\\
		MMR \cite{praveen_kumar_critical_2021} & 0.01 & 0.01 & 0.01 & 0.17 & 0.16 & 0.17\\
		ML \cite{praveen_kumar_critical_2021} & 0.13 & 0.25 & 0.2 & 3.75 & 2.83 & 1.52\\
		NS \cite{praveen_kumar_critical_2021} & 0.02 & 0.02 & 0.03 & 0.15 & 0.16 & 0.22\\
		ST \cite{lin_tactics_2017} & 0.15 & 0.65 & 1.48 & 8.08 & 15.23 & \textbf{22.0}\\
		\bottomrule
	\end{tabular}
\end{table}

\begin{table}
	\centering
	\caption{{Reward of the RIAL systems for the ablation test. Clean reward shown in brackets next to the environment name.}}
	\label{tab:RIAL_overview_reward}
	\begin{tabular}{lccccccccc}\toprule
		  & \multicolumn{3}{l}{{Nav (\textit{0.68})}} & \multicolumn{3}{l}{{PP-O (\textit{38.8})}} & \multicolumn{3}{l}{{PP-D (\textit{35.5})}}\\\cmidrule(lr){2-4} \cmidrule(lr){5-7} \cmidrule(lr){8-10}
		Attack name & 0.25 & 0.5 & 0.75 & 0.25 & 0.5 & 0.75 & 0.25 & 0.5 & 0.75 \\ \midrule
		J-M (Ours) & 1.2 & \textbf{0.23} & \textbf{0.11} & 39.4 & 38.7 & 39.5 & 35.2 & \textbf{33.2} & \textbf{29.7}\\
		J-W (Ours) & 1.16 & \textbf{0.37} & \textbf{0.14} & 38.6 & 38.7 & 39.7 & 35.6 & 34.6 & 32.9\\
		CBTS \cite{zheng_vulnerability_2021} & 1.09 & 0.46 & \textbf{0.08} & 38.7 & 39.0 & 38.2 & 35.5 & 35.5 & 35.2\\
		MMR \cite{praveen_kumar_critical_2021} & 0.74 & \textbf{0.37} & 0.52 & 39.3 & 39.3 & 37.9 & 35.4 & 35.2 & 36.5\\
		ML \cite{praveen_kumar_critical_2021} & \textbf{0.43} & 0.61 & 0.97 & 41.4 & 40.6 & \textbf{36.6} & 35.7 & 35.2 & 34.9\\
		NS \cite{praveen_kumar_critical_2021} & \textbf{0.49} & 0.63 & 0.5 & 40.0 & 38.9 & 40.0 & 35.4 & 35.4 & 36.2\\
		VL \cite{praveen_kumar_critical_2021} & 1.38 & 0.53 & \textbf{0.03} & 39.8 & 39.9 & 38.5 & 34.6 & 35.8 & 36.0\\
		ST \cite{lin_tactics_2017} & 1.36 & 0.51 & \textbf{-0.01} & 40.5 & 39.4 & 39.2 & 35.5 & 35.8 & 36.1\\
		\bottomrule
	\end{tabular}
	\begin{tabular}{lcccccc}\toprule
		  & \multicolumn{3}{l}{{TJ-S (\textit{-1.85})}} & \multicolumn{3}{l}{{TJ-L (\textit{-25.8})}}\\\cmidrule(lr){2-4} \cmidrule(lr){5-7}
		Attack name & 0.25 & 0.5 & 0.75 & 0.25 & 0.5 & 0.75 \\ \midrule
		J-M (Ours) & -1.78 & -1.85 & \textbf{-2.08} & -24.8 & -24.9 & -26.1\\
		J-W (Ours) & -1.72 & -1.85 & \textbf{-2.08} & -25.0 & -25.3 & -27.1\\
		CBTS \cite{zheng_vulnerability_2021} & -1.87 & -1.87 & -1.97 & -25.1 & -26.4 & -25.6\\
		MMR \cite{praveen_kumar_critical_2021} & \textbf{-2.25} & \textbf{-2.04} & -1.78 & -27.0 & \textbf{-28.3} & \textbf{-32.8}\\
		ML \cite{praveen_kumar_critical_2021} & \textbf{-2.16} & \textbf{-2.2} & -1.89 & \textbf{-28.6} & \textbf{-30.1} & -24.7\\
		NS \cite{praveen_kumar_critical_2021} & -1.87 & -1.88 & -1.74 & -26.3 & -24.9 & -27.6\\
		VL \cite{praveen_kumar_critical_2021} & -1.69 & -1.88 & \textbf{-2.1} & -24.8 & -26.0 & -26.5\\
		ST \cite{lin_tactics_2017} & -1.69 & -1.86 & \textbf{-2.08} & -24.6 & -26.0 & -26.7\\
		\bottomrule
	\end{tabular}
\end{table}
\begin{table}
	\centering
	\caption{{Rate of successful action changes against the RIAL systems for the ablation test.}}
	\label{tab:RIAL_overview_action_changed}
	\begin{tabular}{lccccccccc}\toprule
		  & \multicolumn{3}{l}{Nav} & \multicolumn{3}{l}{PP-O} & \multicolumn{3}{l}{PP-D}\\\cmidrule(lr){2-4} \cmidrule(lr){5-7} \cmidrule(lr){8-10}
		Attack name & 0.25 & 0.5 & 0.75 & 0.25 & 0.5 & 0.75 & 0.25 & 0.5 & 0.75 \\ \midrule
		J-M (Ours) & 0.16 & 0.22 & 0.24 & 0.06 & 0.09 & 0.11 & 0.42 & 0.48 & 0.54\\
		J-W (Ours) & 0.21 & 0.31 & 0.3 & 0.06 & 0.08 & 0.09 & 0.16 & 0.37 & 0.45\\
		CBTS \cite{zheng_vulnerability_2021} & 0.07 & 0.09 & 0.11 & 0.0 & 0.0 & 0.01 & 0.03 & 0.1 & 0.15\\
		MMR \cite{praveen_kumar_critical_2021} & 0.18 & 0.25 & 0.28 & 0.05 & 0.07 & 0.07 & 0.22 & 0.29 & 0.32\\
		ML \cite{praveen_kumar_critical_2021} & 0.25 & 0.32 & 0.24 & 0.07 & 0.08 & 0.08 & 0.07 & 0.17 & 0.26\\
		NS \cite{praveen_kumar_critical_2021} & 0.23 & 0.22 & 0.25 & 0.16 & 0.17 & 0.16 & 0.09 & 0.22 & 0.28\\
		VL \cite{praveen_kumar_critical_2021} & 0.17 & 0.23 & 0.28 & 0.08 & 0.09 & 0.08 & 0.23 & 0.3 & 0.33\\
		ST \cite{lin_tactics_2017} & 0.15 & 0.21 & 0.25 & 0.05 & 0.07 & 0.07 & 0.18 & 0.27 & 0.31\\
		\bottomrule
	\end{tabular}
	\begin{tabular}{lcccccc}\toprule
		  & \multicolumn{3}{l}{TJ-S} & \multicolumn{3}{l}{TJ-L}\\\cmidrule(lr){2-4} \cmidrule(lr){5-7}
		Attack name & 0.25 & 0.5 & 0.75 & 0.25 & 0.5 & 0.75 \\ \midrule
		J-M (Ours) & 0.01 & 0.03 & 0.06 & 0.0 & 0.0 & 0.0\\
		J-W (Ours) & 0.01 & 0.03 & 0.05 & 0.0 & 0.0 & 0.02\\
		CBTS \cite{zheng_vulnerability_2021} & 0.02 & 0.02 & 0.03 & 0.0 & 0.0 & 0.0\\
		MMR \cite{praveen_kumar_critical_2021} & 0.56 & 0.5 & 0.61 & 0.08 & 0.0 & 0.0\\
		ML \cite{praveen_kumar_critical_2021} & 0.14 & 0.21 & 0.35 & 0.07 & 0.21 & 0.01\\
		NS \cite{praveen_kumar_critical_2021} & 0.14 & 0.11 & 0.11 & 0.03 & 0.02 & 0.0\\
		VL \cite{praveen_kumar_critical_2021} & 0.0 & 0.0 & 0.0 & 0.0 & 0.0 & 0.01\\
		ST \cite{lin_tactics_2017} & 0.0 & 0.0 & 0.0 & 0.0 & 0.0 & 0.01\\
		\bottomrule
	\end{tabular}
\end{table}
\begin{table}
	\centering
	\caption{{Task metric of the RIAL systems for the ablation test. Clean task metric shown in brackets next to the environment name.}}
	\label{tab:RIAL_overview_t_metric}
	\begin{tabular}{lccccccccc}\toprule
		  & \multicolumn{3}{l}{{Nav (\textit{1.85})}} & \multicolumn{3}{l}{{PP-O (\textit{4.33})}} & \multicolumn{3}{l}{{PP-D (\textit{4.07})}}\\\cmidrule(lr){2-4} \cmidrule(lr){5-7} \cmidrule(lr){8-10}
		Attack name & 0.25 & 0.5 & 0.75 & 0.25 & 0.5 & 0.75 & 0.25 & 0.5 & 0.75 \\ \midrule
		J-M (Ours) & 2.31 & \textbf{1.6} & \textbf{1.51} & 4.35 & 4.34 & 4.43 & 4.12 & 4.1 & 4.08\\
		J-W (Ours) & 2.3 & 1.73 & \textbf{1.53} & 4.26 & 4.31 & 4.45 & 4.12 & 4.09 & 4.11\\
		CBTS \cite{zheng_vulnerability_2021} & 2.16 & 1.85 & \textbf{1.55} & 4.28 & 4.36 & 4.33 & 4.04 & 4.08 & 4.11\\
		MMR \cite{praveen_kumar_critical_2021} & 2.06 & 1.98 & 2.09 & 4.33 & 4.39 & 4.31 & 4.07 & 4.06 & 4.19\\
		ML \cite{praveen_kumar_critical_2021} & 1.79 & 2.0 & 2.41 & 4.5 & 4.5 & 4.24 & 4.07 & 4.09 & 4.07\\
		NS \cite{praveen_kumar_critical_2021} & 1.8 & 1.88 & 1.76 & 4.45 & 4.35 & 4.48 & 4.02 & 4.07 & 4.18\\
		VL \cite{praveen_kumar_critical_2021} & 2.43 & 2.02 & \textbf{1.56} & 4.36 & 4.45 & 4.4 & 3.98 & 4.11 & 4.15\\
		ST \cite{lin_tactics_2017} & 2.41 & 1.95 & \textbf{1.47} & 4.42 & 4.4 & 4.44 & 4.07 & 4.12 & 4.16\\
		\bottomrule
	\end{tabular}
	\begin{tabular}{lcccccc}\toprule
		  & \multicolumn{3}{l}{{TJ-S (\textit{0.0})}} & \multicolumn{3}{l}{{TJ-L (\textit{0.16})}}\\\cmidrule(lr){2-4} \cmidrule(lr){5-7}
		Attack name & 0.25 & 0.5 & 0.75 & 0.25 & 0.5 & 0.75 \\ \midrule
		J-M (Ours) & 0.0 & 0.0 & 0.01 & 0.15 & 0.11 & 0.13\\
		J-W (Ours) & 0.0 & 0.0 & 0.01 & 0.16 & 0.17 & 0.19\\
		CBTS \cite{zheng_vulnerability_2021} & 0.0 & 0.0 & 0.0 & 0.14 & 0.25 & 0.14\\
		MMR \cite{praveen_kumar_critical_2021} & 0.0 & 0.0 & 0.0 & 0.28 & \textbf{0.41} & \textbf{0.67}\\
		ML \cite{praveen_kumar_critical_2021} & 0.0 & 0.0 & 0.0 & \textbf{0.44} & \textbf{0.51} & 0.03\\
		NS \cite{praveen_kumar_critical_2021} & 0.0 & 0.0 & 0.0 & 0.27 & 0.12 & 0.25\\
		VL \cite{praveen_kumar_critical_2021} & 0.0 & 0.0 & 0.0 & 0.13 & 0.18 & 0.13\\
		ST \cite{lin_tactics_2017} & 0.0 & 0.0 & 0.0 & 0.11 & 0.17 & 0.15\\
		\bottomrule
	\end{tabular}
\end{table}

\subsection{Loss and message selection functions}

In our ablation study of the loss and message selection functions, we found that using ranked message selection improved or maintained the effectiveness of the attack against most systems, but different loss functions were more effective against different systems.

We show the reward of the observation sharing system in Table \ref{tab:ObservationSharing_msg_ablation_reward}.
We observe that for the navigation environment, attacks using either ranked message selection or the maximum loss function were effective at moderate attack rates, but at higher attack rate the maximum loss function and random message selection with the untargeted loss function were the most effective.
The effectiveness of the random message selection with the untargeted loss function over the ranked message selection with the untargeted loss function is the only exception to the improvement offered by the ranked message selection function. This exception may be explained by the local optima phenomenon we mentioned earlier.
For the orthogonal PredatorPrey, all attacks are effective at low attack rates but ranked message selection and the untargeted loss function offer improved performance at increasing attack rates.
For the diagonal version, ranked message selection with weighted loss function was the most effective.
For the TrafficJunction environments, the maximum and weighted loss functions were the most effective attacks. For the small environment at low attack rates, ranked message selection with the maximum loss function was required to achieve significantly similar results to the weighted loss function attacks.

We present the results of our ablation test on RIAL in Table \ref{tab:RIAL_msg_ablation_reward}.
In the navigation environment, maximum and weighted loss functions are more effective at moderate attack rates with untargeted loss function requiring the ranked message selection method to match the performance. 
No combination of message selection and loss function made our Jacobian attack effective in the orthogonal PredatorPrey or large TrafficJunction environments. 
The combination of the maximum loss function and ranked message selection was the most effective in the diagonal version, but only at moderate and high attack rates.
Against the small TrafficJunction environment, our attacks were only effective at the highest attack rates.

\begin{table}
	\centering
	\caption{{Reward of the observation sharing systems for the ablation test. Clean reward shown in brackets next to the environment name.}}
	\label{tab:ObservationSharing_msg_ablation_reward}
	\begin{tabular}{lccccccccc}\toprule
		 {Attack} & \multicolumn{3}{l}{{Nav (\textit{1.56})}} & \multicolumn{3}{l}{{PP-O (\textit{42.4})}} & \multicolumn{3}{l}{{PP-D (\textit{35.6})}}\\\cmidrule(lr){2-4} \cmidrule(lr){5-7} \cmidrule(lr){8-10}
		name & 0.25 & 0.5 & 0.75 & 0.25 & 0.5 & 0.75 & 0.25 & 0.5 & 0.75 \\ \midrule
		Rand-M & 2.02 & \textbf{1.4} & \textbf{0.52} & \textbf{40.1} & 37.8 & 34.5 & 30.6 & 26.8 & 23.2\\
		Rand-W & 2.04 & 1.49 & 0.76 & \textbf{40.2} & 38.3 & 34.5 & 25.9 & 24.4 & 20.8\\
		Rand-U & 2.0 & 1.49 & \textbf{0.63} & \textbf{38.9} & \textbf{34.5} & \textbf{30.5} & 33.6 & 31.1 & 27.3\\
		Rank-M & 1.98 & \textbf{1.39} & \textbf{0.48} & \textbf{38.8} & \textbf{34.6} & \textbf{31.4} & 28.6 & 24.0 & 19.4\\
		Rank-W & 1.97 & \textbf{1.46} & 0.7 & \textbf{39.6} & 35.7 & \textbf{31.5} & \textbf{24.0} & \textbf{21.7} & \textbf{17.1}\\
		Rank-U & 1.97 & \textbf{1.42} & 0.7 & \textbf{38.7} & \textbf{34.2} & \textbf{30.3} & 32.8 & 30.6 & 26.7\\
		\bottomrule
	\end{tabular}
	\begin{tabular}{lcccccc}\toprule
		 {Attack} & \multicolumn{3}{l}{{TJ-S (\textit{-1.33})}} & \multicolumn{3}{l}{{TJ-L (\textit{-28.2})}}\\\cmidrule(lr){2-4} \cmidrule(lr){5-7}
		name & 0.25 & 0.5 & 0.75 & 0.25 & 0.5 & 0.75 \\ \midrule
		Rand-M & -2.43 & -7.52 & \textbf{-16.74} & \textbf{-152.9} & \textbf{-227.7} & \textbf{-179.7}\\
		Rand-W & \textbf{-3.03} & \textbf{-7.85} & \textbf{-17.31} & \textbf{-164.2} & \textbf{-219.6} & \textbf{-201.2}\\
		Rand-U & -2.65 & -6.97 & -9.8 & -31.1 & -31.0 & -30.8\\
		Rank-M & \textbf{-3.24} & \textbf{-9.82} & \textbf{-18.3} & \textbf{-152.2} & \textbf{-218.4} & \textbf{-177.6}\\
		Rank-W & \textbf{-3.62} & \textbf{-8.77} & \textbf{-18.09} & \textbf{-169.8} & \textbf{-210.6} & \textbf{-190.4}\\
		Rank-U & -2.82 & -7.48 & -9.49 & -31.1 & -31.7 & -31.2\\
		\bottomrule
	\end{tabular}
\end{table}

\begin{table}
	\centering
	\caption{{Reward of the RIAL systems. Clean reward shown in brackets next to the environment name.}}
	\label{tab:RIAL_msg_ablation_reward}
	\begin{tabular}{lccccccccc}\toprule
		 {Attack} & \multicolumn{3}{l}{{Nav (\textit{0.68})}} & \multicolumn{3}{l}{{PP-O (\textit{38.8})}} & \multicolumn{3}{l}{{PP-D (\textit{35.5})}}\\\cmidrule(lr){2-4} \cmidrule(lr){5-7} \cmidrule(lr){8-10}
		name & 0.25 & 0.5 & 0.75 & 0.25 & 0.5 & 0.75 & 0.25 & 0.5 & 0.75 \\ \midrule
		Rand-M & 1.21 & \textbf{0.25} & \textbf{0.19} & 39.3 & 38.4 & 38.9 & 35.0 & 34.3 & 33.7\\
		Rand-W & 1.15 & \textbf{0.39} & \textbf{0.17} & 39.4 & 39.0 & 39.3 & 36.2 & 34.6 & 34.6\\
		Rand-U & 1.18 & 0.42 & \textbf{0.17} & 38.5 & 38.5 & 38.8 & 35.2 & 35.9 & 35.0\\
		Rank-M & 1.2 & \textbf{0.23} & \textbf{0.11} & 39.4 & 38.7 & 39.5 & 35.2 & \textbf{33.2} & \textbf{29.7}\\
		Rank-W & 1.16 & \textbf{0.37} & \textbf{0.14} & 38.6 & 38.7 & 39.7 & 35.6 & 34.6 & 32.9\\
		Rank-U & 1.15 & \textbf{0.33} & \textbf{0.06} & 38.2 & 38.5 & 39.3 & 35.8 & 35.7 & 34.4\\
		\bottomrule
	\end{tabular}
	\begin{tabular}{lcccccc}\toprule
		 {Attack} & \multicolumn{3}{l}{{TJ-S (\textit{-1.85})}} & \multicolumn{3}{l}{{TJ-L (\textit{-25.8})}}\\\cmidrule(lr){2-4} \cmidrule(lr){5-7}
		name & 0.25 & 0.5 & 0.75 & 0.25 & 0.5 & 0.75 \\ \midrule
		Rand-M & -1.77 & -1.9 & -1.97 & -24.8 & -25.0 & -26.7\\
		Rand-W & -1.71 & -1.84 & \textbf{-2.11} & -25.0 & -24.9 & -26.4\\
		Rand-U & -1.8 & -1.87 & \textbf{-2.04} & -24.0 & -25.7 & -25.6\\
		Rank-M & -1.78 & -1.85 & \textbf{-2.08} & -24.8 & -24.9 & -26.1\\
		Rank-W & -1.72 & -1.85 & \textbf{-2.08} & -25.0 & -25.3 & \textbf{-27.1}\\
		Rank-U & -1.82 & -1.87 & \textbf{-2.13} & -23.6 & -25.8 & -26.0\\
		\bottomrule
	\end{tabular}
\end{table}

\subsection{Limitations}
\label{sec:limitations}

A key assumption in our work is that the system is well trained and accurately approximates the Q-function. However, this assumption is not well supported by the evidence because some attacks improve the performance of the system.

The conclusions we draw from our empirical evaluation are constrained by the communication methods and environments that we used.
Our experiments attack two communication methods and we are unable to make conclusions on the effectiveness of our attack on other communication methods. This constraint also includes additional aspects of communication including network topology, delayed messages, and noisy channels, which may affect the vulnerability of the system.
The other limitation is the set of environments we used. We focused on three environments with diverse tasks and action sets but with relatively small observation spaces. Larger observations may decrease the effectiveness of attacks due to the reduced relative size of the perturbation. We considered two configurations of the PredatorPrey and TrafficJunction environments to test the impact of more actions and more agents respectively. However, our conclusions may not hold for other configurations of these environments. It would be valuable for future work to evaluate additional communication algorithms and environments.

The effectiveness of using thresholds derived from the collected data from clean episodes to control the attack rate is not as high as expected.
We believe this is due to attacks altering the dynamics of the system and thus changing the distribution of values used by the tempo functions. Because the performance of the attacks is controlled by the threshold, a feedback loop is introduced to the system making exact control of the attack rate very difficult. Our binning method allows us to control for the variation in attack rate and enables a fair comparison between methods, with high computational efficiency. However, a better but more computationally expensive approach may identify appropriate thresholds by running the attacks. This approach may also yield data that could be valuable in developing a deeper understanding of tempo functions.

Our study focuses on the vulnerability of undefended agents, without considering the impact of potential defences. Communication algorithms that use a discrete set of messages, which would include both observation sharing in environments with a discrete set of observations and RIAL in all environments, could be defended with a simple input correction defence that returns messages that deviate from the message set to the nearest valid message. This would remove small perturbations and constrain the attacker to message replacement attacks. However, the effectiveness of such a defence would be determined by properties of the message set, namely, the number of messages and the distance between messages. Future work could also consider the effectiveness of other defences including adversarial training \cite{goodfellow_explaining_2015, madry_towards_2018, tu_adversarial_2021} and how these defences alter the effectiveness of different attacks.

\section{Conclusion}

Our study enhances the effectiveness of single-victim communication perturbation attacks, providing new insights into the vulnerability of multi-agent systems.
Our findings reveal that identifying and exploiting the weakest links in a multi-agent system by tailoring the loss function and selecting a suitable attack timing, victim, and message significantly improves attack effectiveness and affects systems that would otherwise appear to be robust to attack. Moreover, our results provide new insight into how the properties of the communication method and environment affect the robustness of cooperative MARL systems under adversarial conditions. 
We develop a framework for modelling single-victim communication perturbation attacks, revealing key features of these attacks that have previously been unexplored in this context, namely, when-to-attack, who-to-attack, and where-to-attack. We address these overlooked aspects and find that our methods are able to achieve a greater impact in a wider range of scenarios including those where previous attacks had no impact. In considering when-to-attack we present a new technique that exploits an adversary's white-box knowledge of the victim to select an effective attack tempo. Who-to-attack has not been previously addressed in AML attacks on MARL, and we demonstrate that tempo functions can be extended to also select the weakest link in the system,  allowing the adversary to achieve highly effective attacks. Similarly, where-to-attack has not been previously addressed for communication perturbation attacks and is addressed by our novel method of ranked message selection. We empirically demonstrate that our method enables attacks to achieve a similar or greater impact than random message selection in 29 of the 30 tested scenarios. Finally, we propose two new loss functions that address flaws in the basic untargeted loss function. We empirically demonstrate the general effectiveness and specific strengths of these loss functions.

\newpage
\FloatBarrier
\bibliographystyle{ieeetr}
\bibliography{references}

\end{document}